\documentclass{article}
\PassOptionsToPackage{sort,square,comma,numbers}{natbib}
\usepackage[final]{nips_2016}

\usepackage[utf8]{inputenc} 
\usepackage[T1]{fontenc}    
\usepackage{hyperref}       
\usepackage{url}            
\usepackage{booktabs}       
\usepackage{amsfonts}       
\usepackage{nicefrac}       
\usepackage{microtype}      

\usepackage{amsmath}
\usepackage{mathtools}
\usepackage{subcaption}
\usepackage{tikz}
\usepackage{graphicx}
\usepackage{placeins}
\usepackage{bbm}
\usepackage{bm}
\usepackage{color}
\usepackage{multirow}
\usepackage{caption}
\usepackage{listings}
\usepackage[lighttt]{lmodern}
\usepackage{framed}

\let\llncssubparagraph\subparagraph
\let\subparagraph\paragraph
\usepackage[compact]{titlesec}
\let\subparagraph\llncssubparagraph

\lstdefinelanguage{neulang}
{
  morekeywords=[2]{STOP,ZERO,INC,ADD,SUB,DEC,MIN,MAX,READ,WRITE,JEZ},
  morekeywords=[3]{var},
  sensitive=true, 
}

\titlespacing\section{0pt}{9pt plus 4pt minus 2pt}{4pt plus 2pt minus 2pt}
\titlespacing\subsection{0pt}{9pt plus 4pt minus 2pt}{2pt plus 2pt minus 2pt}
\titlespacing\subsubsection{0pt}{9pt plus 4pt minus 2pt}{1pt plus 2pt minus 2pt}

\title{Adaptive Neural Compilation}

\author{
  \begin{tabular}[t]{cc}
Rudy Bunel\thanks{The first two authors contributed equally.}  & Alban Desmaison\footnotemark[1]  \\
        \normalfont University of Oxford & \normalfont University of Oxford \\
        \normalfont \texttt{rudy@robots.ox.ac.uk} & \normalfont \texttt{alban@robots.ox.ac.uk} \\ \\
\end{tabular}
\\
\begin{tabular}[t]{ccc}
\bf Pushmeet Kohli  &\bf Philip H.S. Torr & \bf M. Pawan Kumar \\
        Microsoft Research & University of Oxford & University of Oxford \\
        \texttt{pkohli@microsoft.com} & \texttt{philip.torr@eng.ox.ac.uk} & \texttt{pawan@robots.ox.ac.uk}
\end{tabular}
}

\makeatletter
\@addtoreset{footnote}{section}
\makeatother

\begin{document}

\maketitle

\begin{abstract}
This paper proposes an adaptive neural-compilation framework to address the problem of efficient program learning.
Traditional code optimisation strategies used in compilers are based on applying pre-specified set of transformations that make the code faster to execute without changing its semantics.
In contrast, our work involves adapting programs to make them more efficient while considering correctness only on a target input distribution.
Our approach is inspired by the recent works on differentiable representations of programs.
We show that it is possible to compile programs written in a low-level  language to a differentiable representation.
We also show how programs in this representation can be optimised to make them efficient on a target distribution of inputs.
Experimental results demonstrate that our approach enables learning \emph{specifically-tuned} algorithms for given data distributions with a high success rate.
\end{abstract}

\section{Introduction}
\label{sec:intro}
Algorithm design often requires making simplifying assumptions about the input data.
Consider, for instance, the computational problem of accessing an element in a linked list.
Without the knowledge of the input data distribution, one can only specify an algorithm that runs in a time linear in the number of elements of the list.
However, suppose all the linked lists that we encountered in practice were ordered in memory. Then it would be advantageous to design an algorithm specifically for this task as it can lead to a constant running time.
Unfortunately, the input data distribution of a real world problem cannot be easily specified as in the above simple example.
The best that one can hope for is to obtain samples drawn from the distribution. A natural question that arises
from these observations: ``How can we adapt a generic algorithm for a computational task using samples
from an unknown input data distribution?''

The process of finding the most efficient implementation of an algorithm has received considerable attention in the theoretical computer science and code optimisation community.
Recently, Conditionally Correct Superoptimization~\citep{sharma2015conditionally} was proposed as a method for leveraging samples of the input data distribution to go beyond semantically equivalent optimisation and towards data-specific performance improvements.
The underlying procedure is based on a stochastic search over the space of all possible programs.
Additionally, they restrict their applications to reasonably small, loop-free programs, thereby limiting their impact in practice.

In this work, we take inspiration from the recent wave of machine-learning frameworks for estimating programs.
Using recurrent models, \citet{graves2014neural} introduced a fully differentiable representation of a program, enabling the use of gradient-based methods to learn a program from examples.
Many other models have been published recently \citep{nram,kaiser2015neural,joulin2015inferring,grefenstette2015learning} that build and improve on the early work by \citet{graves2014neural}.
Unfortunately, these models are usually complex to train and need to rely on methods such as curriculum learning or gradient noise to reach good solutions as shown by~\citet{gradient-noise}.
Moreover, their interpretability is limited.
The learnt model is too complex for the underlying algorithm to be recovered and transformed into a regular computer program.

The main focus of the machine-learning community has thus far been on learning programs from scratch,
with little emphasis on running time.
 However, for nearly all computational problems,
it is feasible to design generic algorithms for the worst-case.
We argue that a more pragmatic goal for the machine learning community is to design methods for adapting existing programs for specific input data distributions.
To this end, we propose the Adaptive Neural Compiler (ANC).
We design a compiler capable of mechanically converting algorithms to a differentiable representation, thereby providing adequate initialisation to the difficult problem of optimal program learning.
We then present a method to improve this compiled program using data-driven optimisation, alleviating the need to perform a wide search over the set of all possible programs.
We show experimentally that this framework is capable of adapting simple generic algorithms to perform better on given datasets.

\section{Related Works}
\label{sec:related-works}
The idea of compiling programs to neural networks has previously been explored in the literature.
\citet{siegelmann1994neural} described how to build a Neural Network that would perform the same operations as a given program.
A compiler has been designed by \citet{gruau1995neural} targeting an extended version of Pascal.
A complete implementation was achieved when \citet{neto2003symbolic} wrote a compiler for NETDEF, a language based on the Occam programming language.
While these methods allow us to obtain an exact representation of a program as a neural network, they do not lend themselves to optimisation to improve the original program.
Indeed, in their formulation, each elementary step of a program is expressed as a group of neurons with a precise topology, set of weights and biases, thereby rendering learning via gradient descent infeasible.
Performing gradient descent in this parameter space would result in invalid operations and thus is unlikely to lead to any improvement.
The recent work by \citet{npi} on Neural Programmer-Interpreters (NPI) can also be seen as a way to compile any program into a neural network by learning a model that mimic the program.
While more flexible than the previous approaches, the NPI is unable to improve on a learned program due to its dependency on a non-differentiable environment.

Another approach to this learning problem is the one taken by the code optimisation community.
By exploring the space of all possible programs, either exhaustively~\citep{massalin1987superoptimizer} or in a stochastic manner~\citep{schkufza2013stochastic}, they search for programs having the same results but being more efficient.
The work of \citet{sharma2015conditionally} broadens the space of acceptable improvements to data-specific optimisations as opposed to the provably equivalent transformations that were previously the only ones considered.
However, this method is still reliant on non-gradient-based methods for efficient exploration of the space.
By representing everything in a differentiable manner, we aim to obtain gradients to guide the exploration.

Recently, \citet{graves2014neural} introduced a learnable representation  of programs, called the Neural Turing Machine (NTM).
The NTM uses an LSTM as a Controller, which outputs commands to be executed by a deterministic differentiable Machine.
From examples of input/output sequences, they manage to learn a Controller such that the model becomes capable of performing simple algorithmic tasks.
Extensions of this model have been proposed in \citep{joulin2015inferring,grefenstette2015learning} where the memory tape was replaced by differentiable versions of stacks or lists.
\citet{nram} modified the NTM to introduce a notion of pointers making it more amenable to represent traditional programs.
Parallel works have been using Reinforcement Learning techniques such as the REINFORCE algorithm \citep{williams1992simple,andrychowicz2016learning,zaremba2015reinforcement} or Q-learning \citep{zaremba2015learning} to be able to work with non differentiable versions of the above mentioned models.
All these models are trained only with a loss based on the difference between the output of the model and the expected output.
This weak supervision leads to a complex training.
For instance the Neural RAM \citep{nram} requires a high number of random restarts before converging to a correct solution~\citep{gradient-noise}, even when using the best hyperparameters obtained through a large grid search.

In our work, we will first show that we can design a new neural compiler whose target will be a Controller-Machine model.
This makes the compiled model amenable to learning from examples.
Moreover, we can use it as initialisation for the learning procedure, allowing us to aim for the more complex task of finding an efficient algorithm.

\section{Model}
\label{sec:model}
Our model is composed of two parts: \textit{(i)} a Controller, in charge of specifying what should be executed;
and \textit{(ii)} a Machine, following the commands of the Controller.
We start by describing the global architecture of the model.
For the sake of simplicity, the general description will present a non-differentiable version of the model.
Section~\ref{subsec:diff} will then explain the modifications required to make this model completely differentiable.
A more detailed description of the model is provided in appendix~\ref{sec:supp-detailed-model}.

\subsection{General Model}
\label{subsec:gen_model}
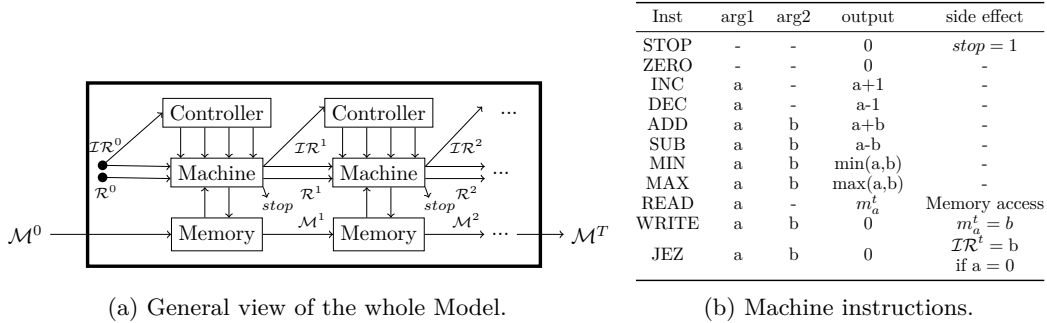
\begin{figure}
\captionsetup{font=footnotesize}
\begin{subfigure}[b]{0.6\textwidth}
  \begin{center}
    \usetikzlibrary{decorations.pathreplacing,calc,matrix,circuits.logic.US,positioning,matrix, fit}
\resizebox{\textwidth}{!}{
    \begin{tikzpicture}

      {
        \node[draw] (controller-t) {Controller};
        \node[draw, below = 0.5cm of controller-t] (machine-t) {Machine};
        \node[draw, below = 0.5cm of machine-t] (memory-t) {Memory};
        \draw[->] ([xshift=-0.6cm]controller-t.south) to ([xshift=-0.6cm]machine-t.north);
        \draw[->] ([xshift=-0.2cm]controller-t.south) to ([xshift=-0.2cm]machine-t.north);
        \draw[->] ([xshift=0.6cm]controller-t.south) to ([xshift=0.6cm]machine-t.north);
        \draw[->] ([xshift=0.2cm]controller-t.south) to ([xshift=0.2cm]machine-t.north);
        \draw[<-] ([xshift=-0.2cm]machine-t.south) to ([xshift=-0.2cm]memory-t.north);
        \draw[->] ([xshift=0.2cm]machine-t.south) to ([xshift=0.2cm]memory-t.north);
        \node[fit=(controller-t)(memory-t)(machine-t)](time-t){};

        \node[draw, right = 25pt of controller-t] (controller-t+1) {Controller};
        \node[draw, below = 0.5cm of controller-t+1] (machine-t+1) {Machine};
        \node[draw, below = 0.5cm of machine-t+1] (memory-t+1) {Memory};
        \draw[->] ([xshift=-0.6cm]controller-t+1.south) to ([xshift=-0.6cm]machine-t+1.north);
        \draw[->] ([xshift=-0.2cm]controller-t+1.south) to ([xshift=-0.2cm]machine-t+1.north);
        \draw[->] ([xshift=0.6cm]controller-t+1.south) to ([xshift=0.6cm]machine-t+1.north);
        \draw[->] ([xshift=0.2cm]controller-t+1.south) to ([xshift=0.2cm]machine-t+1.north);
        \draw[<-] ([xshift=-0.2cm]machine-t+1.south) to ([xshift=-0.2cm]memory-t+1.north);
        \draw[->] ([xshift=0.2cm]machine-t+1.south) to ([xshift=0.2cm]memory-t+1.north);
        \node[fit=(controller-t+1)(memory-t+1)(machine-t+1)](time-t+1){};

        \draw[->] ([yshift=-0.1cm]machine-t.east) to node[pos=0.7, below]{\scriptsize $\mathcal{R}^1$} ([yshift=-0.1cm]machine-t+1.west);
        \draw[->] ([yshift=0.1cm]machine-t.east) to node[pos=0.7, above, yshift=3pt]{\scriptsize $\mathcal{IR}^1$} ([yshift=0.1cm]machine-t+1.west);
        \draw[->] ([yshift=0.1cm]machine-t.east) to node[midway, above, sloped]{} ([yshift=0.1cm]controller-t+1.west);
        \draw[->] (memory-t) to node[pos=0.7, above]{\scriptsize $\mathcal{M}^{1}$}(memory-t+1);

        \node[below right = 0.1cm and -0.15cm of machine-t] (stop-t) {\scriptsize $stop$};
        \draw[->] (machine-t.south east) to ([yshift=-2pt, xshift=-5pt]stop-t.north);
        \node[below right = 0.1cm and -0.15cm of machine-t+1] (stop-t+1) {\scriptsize $stop$};
        \draw[->] (machine-t+1.south east) to ([yshift=-2pt, xshift=-5pt]stop-t+1.north);

        \node[above left = -0.45cm and of machine-t] (ini-reg) {};
        \fill (ini-reg) circle [radius=0.08cm];
        \node[above left = -0.25cm and of machine-t] (ini-regI) {};
        \fill (ini-regI) circle [radius=0.08cm];
        \draw[->] (ini-reg.center) to node[pos=0.05, below]{\scriptsize $\mathcal{R}^0$} ([yshift=-0.1cm]machine-t.west);
        \draw[->] (ini-regI.center) to node[pos=0.05, above, yshift=3pt]{\scriptsize $\mathcal{IR}^0$} ([yshift=0.1cm]machine-t.west);
        \draw[->] (ini-regI.center) to node[midway, above, sloped]{} (controller-t.west);

        \node[right = 28pt of controller-t+1] (rec-controller) {...};
        \node[right = 28pt of machine-t+1] (rec-machine) {...};
        \node[right = 28pt of memory-t+1] (rec-mem) {...};
        \node[fit=(rec-controller)(rec-machine)(rec-mem)](time-rec){};

        \draw[->] ([yshift=-0.1cm]machine-t+1.east) to node[pos=0.7, below]{\scriptsize $\mathcal{R}^2$} ([yshift=-0.1cm]rec-machine.west);
        \draw[->] ([yshift=0.1cm]machine-t+1.east) to node[pos=0.7, above, yshift=3pt]{\scriptsize $\mathcal{IR}^2$} ([yshift=0.1cm]rec-machine.west);
        \draw[->] ([yshift=0.1cm]machine-t+1.east) to node[midway, above, sloped]{} ([yshift=0.1cm, xshift=-5pt]rec-controller.west);
        \draw[->] (memory-t+1) to node[pos=0.7, above]{\scriptsize $\mathcal{M}^{2}$}(rec-mem);

        \node[fit=(time-t+1)(time-t)(ini-reg)(time-rec)] (repeated){};
        \draw[ultra thick]
        (repeated.north west) -- (repeated.south west) -- (repeated.south east) -- (repeated.north east) -- cycle;
      }

      {
        \node[draw=none, left = 2cm of memory-t] (memory-0) {$\mathcal{M}^{0}$};
        \draw[->] (memory-0.east) to (memory-t.west);
      }

      {
        \node[draw=none, right = 2.3cm of memory-t+1] (memory-final) {$\mathcal{M}^{T}$};
        \draw[->] (rec-mem.east) to (memory-final.west);

      }

        \end{tikzpicture}
}
    \end{center}
    \caption{General view of the whole Model.}
    \label{fig:gen_view}
\end{subfigure}
\begin{subfigure}[b]{0.4\textwidth}
 \resizebox{\textwidth}{!}{
        \centering
        \begin{tabular}{@{}ccccc@{}}
        \toprule
        Inst & arg1 & arg2 & output & side effect \\
        \midrule
        STOP & - & - & 0 & $stop = 1$\\
        ZERO & - & - & 0 & - \\
        INC & a & - & a+1 & - \\
        DEC & a & - & a-1 & - \\
        ADD & a & b & a+b & - \\
        SUB & a & b & a-b & - \\
        MIN & a & b & min(a,b) & - \\
        MAX & a & b & max(a,b) & - \\
        READ & a & - & $m^t_a$ & Memory access \\
       WRITE & a & b & 0 & $m^t_a = b$\\
       JEZ & a & b & 0 &  \begin{tabular}{@{}c@{}}$\mathcal{IR}^{t}=\text{b}$ \\ if $\text{a}=0$\end{tabular}\\
        \bottomrule
        \end{tabular}
}
    \caption{Machine instructions.}
    \label{tab:instr_list}
\end{subfigure}

  \caption{Model components.}
  \label{fig:full_mod_descr}
\end{figure}

We first define for each timestep $t$ the memory tape that contains $M$ integer values \mbox{$\mathcal{M}^{t} = \{m_{1}^{t}, m_{2}^{t},\dots, m_{M}^{t}\}$}, the registers that contain $R$ values \mbox{$\mathcal{R}^{t} = \{r_{1}^{t}, r_{2}^{t},\dots, r_{R}^{t}\}$} and the instruction register that contain a single value $\mathcal{IR}^{t}$.
We also define a set of instructions that can be executed, whose main role is to perform computations using the registers.
For example, add the values contained in two registers.
We also define as a side effect any action that involves elements other than the input and output values of the instruction.
Interaction with the memory is an example of such side effect.
All the instructions, their computations and side effects are detailed in Figure~\ref{tab:instr_list}.

As can be seen in Figure~\ref{fig:gen_view} the execution model takes as input an initial memory tape $\mathcal{M}^{0}$ and outputs a final memory tape $\mathcal{M}^{T}$ after $T$ steps.
At each step $t$, the Controller uses the instruction register $\mathcal{IR}^{t}$ to compute the command for the Machine.
The command is a 4-tuple ${e, a, b, o}$.
The first element $e$ is the instruction that should be executed by the Machine, enumerated as an integer.
The elements $a$ and $b$ specify which registers should be used as arguments for the given instruction.
The last element $o$ specifies in which register the output of the instruction should be written.
For example, the command $\{ADD, 2, 3, 1\}$ means that only the value of the first register should change, following \mbox{$r^{t+1}_1 = ADD(r^t_2, r^t_3)$}.
Then the Machine will execute this command, updating the values of the memory, the registers and the instruction register.
The Machine always performs two other operations apart from the required instruction.
It outputs a $stop$ flag that allows the model to decide when to stop the execution.
It also increments the instruction register $\mathcal{IR}^{t}$ by one at each iteration.

\subsection{Differentiability}
\label{subsec:diff}
The model presented above is a simple execution machine but it is not differentiable.
In order to be able to train this model end-to-end from a loss defined over the final memory tape, we need to make every intermediate operation differentiable.

To achieve this, we replace every discrete value in our model by a multinomial distribution over all the possible values that could have been taken.
Moreover, each hard choice that would have been non-differentiable is replaced by a continuous soft choice.
We will henceforth use bold letters to indicate the probabilistic version of a value.

First, the memory tape $\mathcal{M}^{t}$ is replaced by an $M \times M$ matrix $\mathbf{M}^{t}$, where $\mathbf{M}^{t}_{i,j}$ corresponds to the probability of $m_{i}^{t}$ taking the value $j$.
The same change is applied to the registers $\mathcal{R}^{t}$, replacing them with an $R \times M$ matrix $\mathbf{R}^{t}$, where $\mathbf{R}^{t}_{i,j}$ represents the probability of $r_{i}^{t}$ taking the value $j$.
Finally, the instruction register is also transformed from a single value $\mathcal{IR}^{t}$ to a vector of size $M$ noted $\bm{\mathcal{IR}}^{t}$, where the $i$-th element represents its probability to take the value $i$.

The Machine does not contain any learnable parameter and will just execute a given command.
To make it differentiable, the Machine now takes as input four probability distributions $\mathbf{e}^t$, $\mathbf{a}^t$, $\mathbf{b}^t$ and $\mathbf{o}^t$, where $\mathbf{e}^t$ is a distribution over instructions, and $\mathbf{a}^t, \mathbf{b}^t \text{ and } \mathbf{o}^t$ are distributions over the registers.
We compute the argument values $\mathbf{arg_{1}}^{t}$ and $\mathbf{arg_{2}}^{t}$ as convex combinations of the different registers:
\begin{equation}
    \mathbf{arg_{1}}^{t} = \sum_{i=1}^{R} \mathbf{a}^{t}_{i} \mathbf{r}^{t}_{i} \qquad \mathbf{arg_{2}}^{t} = \sum_{i=1}^{R} \mathbf{b}^{t}_{i} \mathbf{r}^{t}_{i},
\end{equation}
where $\mathbf{a}^{t}_{i}$ and $\mathbf{b}^{t}_{i}$ are the $i$-th values of the vectors $\mathbf{a}^{t}$ and $\mathbf{b}^{t}$.
Using these values, we can compute the output value of each instruction $k$ using the following formula:
\begin{equation}
  \forall 0 \leq c \leq M \quad \mathbf{out}_{k,c}^{t} = \sum_{0 \leq i,j \leq M} \mathbf{arg_{1}}^{t}_i \cdot \mathbf{arg_{2}}^{t}_j \cdot \mathbbm{1}[g_k(i,j)=c \mod M],
\end{equation}
where $g_k$ is the function associated to the $k$-th instruction as presented in Table~\ref{tab:instr_list}.
Since the executed instruction is controlled by the probability $\mathbf{e}$, the output written to the register will also be a convex combination: \mbox{$\mathbf{out}^t = \sum_{k=1}^{N} \mathbf{e}^{t}_{k} \mathbf{out}_k^t$}, where $N$ is the number of instructions.
This value is then stored into the registers by performing a soft-write parametrised by $\mathbf{o}^{t}$.

A special case is associated with the $stop$ signal.
When executing the model, we keep track of the probability that the program should have terminated before this iteration based on the probability associated at each iteration with the specific instruction that controls this flag.
Once this probability goes over a threshold $\eta_{\text{stop}} \in (0,1]$, the execution is halted.
We applied the same techniques to make the side-effects differentiable, this is presented in appendix~\ref{subsec:side-eff}.

The Controller is the only learnable part of our model.
The first learnable part is the initial values for the registers $\mathbf{R}^{0}$ and for the instruction register $\bm{\mathcal{IR}}^{0}$.
The second learnable part is the parameters of the Controller which computes the required distributions using:
\begin{equation}
\mathbf{e}^t = \mathbf{W}_{e} * \bm{\mathcal{IR}}^{t}, \qquad
\mathbf{a}^t = \mathbf{W}_{a} * \bm{\mathcal{IR}}^{t}, \qquad
\mathbf{b}^t = \mathbf{W}_{b} * \bm{\mathcal{IR}}^{t}, \qquad
\mathbf{o}^t = \mathbf{W}_{o} * \bm{\mathcal{IR}}^{t}
\end{equation}
where $\mathbf{W}_{e}$ is an $N \times M$ matrix and $\mathbf{W}_{a}$, $\mathbf{W}_{b}$ and $\mathbf{W}_{o}$ are $R \times M$ matrices.
A representation of these matrices can be found in Figure~\ref{lst:weight-program}.
The Controller as defined above is composed of four independent, fully-connected layers.
In Section~\ref{subsec:nc} we will see that this complexity is sufficient for our model to be able to represent \emph{any} program.

Henceforth, we will denote by $\bm{\theta} = \{\mathbf{R}^{0}, \bm{\mathcal{IR}}^{0}, \mathbf{W}_{e}, \mathbf{W}_{a}, \mathbf{W}_{b}, \mathbf{W}_{o}\}$ the set of all the learnable parameters of this model.

\section{Adaptative Neural Compiler}
\label{sec:anc}
We will now present the Adaptive Neural Compiler.
Its goal is to find the best set of weights $\bm{\theta}^*$ for a given dataset such that our model will perform the correct input/output mapping as efficiently as it can.
We begin by describing our learning objective in details.
The two subsequent sections will focus on making the optimisation of our learning objective computationally feasible.

\subsection{Objective function}
Our goal is to solve a given algorithmic problem efficiently.
The algorithmic problem is defined as a set of input/output pairs.
We also have access to a generic program that is able to perform the required mapping.
In our example of accessing elements in a linked list, the transformation would consist in writing down the desired value at the specified position in the tape.
The program given to us would iteratively go through the elements of the linked list, find the desired value and write it down at the desired position.
If there exists some bias that would allow this traversal to be faster, we expect the program to exploit it.

Our approach to this problem is to construct a differentiable objective function, mapping controller parameters to a loss.
We define this loss based on the states of the memory tape and outputs of the Controller at each step of the execution.
The precise mathematical formulation for each term of the loss is given in appendix~\ref{sec:supp-loss}.
Here we present the motivation behind each of them.

\paragraph{Correctness}
For a given input, we have the expected output.
We compare the values of the expected output with the final memory tape provided by the execution.

\paragraph{Halting}
To prevent programs to take an infinite amount of time without stopping, we defined a maximum number of iterations $T_{max}$ after which the execution is halted.
Moreover, we add a penalty in the loss if the Controller didn't halt before this limit.

\paragraph{Efficiency}
We penalise each iteration taken by the program where it does not stop.

\paragraph{Confidence}
We add a term which will penalise probability of stopping if the current state of the memory is not the expected one.

If only the correctness term was considered, nothing would encourage the learnt algorithm to halt as soon as it finished. If only correctness and halting were considered, then the program may not halt as early as possible.
Confidence enables the algorithm to evaluate better when to stop.

The loss is a weighted sum of the four above-mentioned terms.
We denote the loss of the $i$-th training sample, given parameters $\bm{\theta}$, as $L_i(\bm{\theta})$.
Our learning objective is then specified as:
\begin{equation}
    \min_{\bm{\theta}} \quad \sum_i L_i(\bm{\theta}) \quad \text{s.t.} \ \bm{\theta} \in \bm{\Theta},
\end{equation}
where $\Theta$ is a set over the parameters such that the outputs of the Controller, the initial values of each register and of the instruction register are all probability distributions.

The above optimisation is a highly non-convex problem.
To be able to solve it using standard gradient descent based methods, we will first need to transform it to an unconstrained problem.
We also know that the result of the optimisation of a non-convex objective function is strongly dependent on the initialisation point.
In the rest of this section, we will first present a small modification to the model that will remove the constraints.
We will then present our Neural Compiler that will provide a good initialisation to solve this problem.

\subsection{Reformulation}
In order to use gradient descent methods without having to project the parameters on $\bm{\Theta}$, we alter the formulation of the controller.
We add a softmax layer after each linear layer ensuring that the constraints on the Controller's output will be respected.
We also apply a softmax to the initial values of the registers and the instructions register, ensuring they will also respect the original constraints.
This way, we transform the constrained-optimisation problem into an unconstrained one, allowing us to use standard gradient descent methods.
As discussed in other works \citep{gradient-noise}, this kind of model is hard to train and requires a high number of random restarts before converging to a good solution.
We will now present a Neural Compiler that will provide good initialisations to help with this problem.

\subsection{Neural Compiler}
\label{subsec:nc}
\begin{figure}[htbp]
\captionsetup{font=footnotesize}
\resizebox{0.3\textwidth}{!}{
\begin{subfigure}[b]{0.38\textwidth}
    \begin{equation*}
      \begin{split}
        &\text{\textbf{var} head} = 0;\\
        &\text{\textbf{var} nb\_jump} = 1;\\
        &\text{\textbf{var} out\_write} = 2;\\
\ \\
        &\text{nb\_jump = \textbf{READ}(nb\_jump);}\\
        &\text{out\_write = \textbf{READ}(out\_write);}\\
        loop:\quad& \text{head = \textbf{READ}(head);}\\
        &\text{nb\_jump = \textbf{DEC}(nb\_jump);}\\
        &\text{\textbf{JEZ}(nb\_jump, }end\text{);}\\
        &\text{\textbf{JEZ}(0, }loop\text{);}\\
        end:\quad&\text{head = \textbf{INC}(head);}\\
        &\text{head = \textbf{READ}(head);}\\
        &\text{\textbf{WRITE}(out\_write, head);}\\
        &\text{\textbf{STOP}();}
      \end{split}
    \end{equation*}
    \caption[]{\label{lst:input-program}Input program}
  \end{subfigure}
}
\hfill
{
\resizebox{0.3\textwidth}{!}{
   \begin{subfigure}[b]{0.38\textwidth}
    Initial Registers:
\begin{flalign*}
    R_{1}&=6;\quad R_{2}=2;\quad R_{3}=0; &\\
    R_{4}&=2;\quad R_{5}=1;\quad R_{6}=0; &\\
    R_{7}&=0;
\vspace{5pt}
\end{flalign*}

    Program:
\begin{flalign*}
    0: R_{5} &= \textbf{READ }(R_{5}, R_{7})&\\
    1: R_{4} &= \textbf{READ }(R_{4}, R_{7})&\\
    2: R_{6} &= \textbf{READ }(R_{6}, R_{7})&\\
    3: R_{5} &= \textbf{DEC}\ \ \ \ (R_{5}, R_{7})&\\
    4: R_{7} &= \textbf{JEZ}\ \ \ \ \ (R_{5}, R_{1})&\\
    5: R_{3} &= \textbf{JEZ}\ \ \ \ \ (R_{3}, R_{2})&\\
    6: R_{6} &= \textbf{INC}\ \ \ \ \ (R_{6}, R_{7})&\\
    7: R_{6} &= \textbf{READ}\ (R_{6}, R_{7})&\\
    8: R_{7} &= \textbf{WRITE}(R_{4}, R_{6})&\\
    9: R_{7} &= \textbf{STOP }\ \ (R_{7}, R_{7})
    \end{flalign*}
    \caption[]{\label{lst:inter-program}Intermediary representation}
  \end{subfigure}
}
}
{
\resizebox{0.32\textwidth}{!}{
  \begin{subfigure}[b]{0.4\textwidth}
    \setcounter{subfigure}{0}
    \renewcommand\thesubfigure{\roman{subfigure}}
\begin{tabular}{cc}
    \begin{subfigure}[b]{0.4\textwidth}
      \centerline{\frame{\includegraphics[height=0.1\textheight]{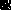}}}
      \caption[]{\label{fig:instr-weight}Instr.}
    \end{subfigure}
 &
    \begin{subfigure}[b]{0.4\textwidth}
      \centerline{\frame{\includegraphics[height=0.1\textheight]{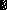}}}
      \caption[]{\label{fig:arg1-weight}Arg1}
    \end{subfigure}
\\
    \begin{subfigure}[b]{0.4\textwidth}
      \centerline{\frame{\includegraphics[height=0.1\textheight]{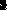}}}
      \caption[]{\label{fig:arg2-weight}Arg2}
    \end{subfigure}
 &
    \begin{subfigure}[b]{0.4\textwidth}
      \centerline{\frame{\includegraphics[height=0.1\textheight]{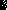}}}
      \caption[]{\label{fig:out-weight}Out}
    \end{subfigure}
\\
\end{tabular}
    \setcounter{subfigure}{2}
    \renewcommand\thesubfigure{\alph{subfigure}}
    \caption[]{\label{lst:weight-program} Weights}
  \end{subfigure}
}
}
  \caption[]{\label{fig:compilation-process}Example of the compilation process.
    (\ref{lst:input-program}) Program written to perform the \textbf{ListK} task. Given a pointer to the head of a linked list, an integer $k$, a target cell and a linked list, write in the target cell the $k$-th element of the list.
    (\ref{lst:inter-program}) Intermediary representation of the program. This corresponds to the instruction that a Random Access Machine would need to perform to execute the program.
    (\ref{lst:weight-program}) Representation of the weights that encodes the intermediary representation. Each row of the matrix correspond to one state/line. Initial value of the registers are also parameters of the model, omitted here.}
\end{figure}

The goal for the Neural Compiler is to convert an algorithm, written as an unambiguous program, to a set of parameters.
These parameters, when put into the controller, will reproduce the exact steps of the algorithm.
This is very similar to the problem framed by \citet{npi}, but we show here a way to accomplish it \emph{without any learning}.

The different steps of the compilation are illustrated in Figure~\ref{fig:compilation-process}.
The first step is to go from the written version of the program to the equivalent list of low level instruction.
This step can be seen as going from Figure~\ref{lst:input-program} to Figure~\ref{lst:inter-program}.
The illustrative example uses a fairly low-level language but traditional features of programming languages such as \texttt{loop}s or \texttt{if}-statements can be supported using the \texttt{JEZ} instruction.
The use of constants as arguments or as values is handled by introducing new registers that hold these values.
The value required to be passed as target position to the \texttt{JEZ} instruction can be resolved at compile time.

Having obtained this intermediate representation, generating the parameters is straightforward.
As can be seen in Figure~\ref{lst:inter-program}, each line contains one instruction, the two input registers and the output register,
and corresponds to a command that the Controller will have to output.
If we ensure that $\bm{\mathcal{IR}}$ is a Dirac-delta distribution on a given value, then the matrix-vector product is equivalent to selecting a row of the weight matrix.
As $\bm{\mathcal{IR}}$ is incremented at each iteration, the Controller outputs the rows of the matrix in order.
We thus have a one-to-one mapping between the lines of the intermediate representation and the rows of the weight matrix.
An example of these matrices can be found in Figure~\ref{lst:weight-program}.
The weight matrix has 10 rows, corresponding to the number of lines of code of our intermediate representation.
On the first line of the matrix corresponding to the first argument~(\ref{lst:weight-program}ii), the fifth element has value $1$, and is linked to the first line of code where the first argument to the READ operation is the fifth register.

The number of rows of the weight matrix is linear in the number of lines of code in the original program.
To output a command, we must be able to index its line with the instruction register $\bm{\mathcal{IR}}$, which means that the largest representable number in our Machine needs to be greater than the number of lines in our program.

Moreover, any program written in a regular assembly language can be rewritten to use only our restricted set of instructions.
This can be done first because all the conditionals of the the assembly language can be expressed as a combination of arithmetic and \texttt{JEZ} instructions.
Secondly because all the arithmetic operations can be represented as a combination of our simple arithmetic operations, \texttt{loop}s and \texttt{if}s statements.
This means that \emph{any} program that can run on a regular computer, can be first rewritten to use our restricted set of instructions and then compiled down to a set of weights for our model.
Even though other models use LSTM as controller, we showed here that a Controller composed of simple linear functions is expressive enough.
The advantage of this simpler model is that we can now easily interpret the weights of our model in a way that would not have be possible if we had a recurrent network as a controller.

The most straightforward way to leverage the results of the compilation is to initialise the Controller with the weights obtained through compilation of the generic algorithm.
To account for the extra softmax layer, we need to multiply the weights produced by the compiler by a large constant to output Dirac-delta distributions.
Some results associated with this technique can be found in Section~\ref{subsec:generalisation-test}.
However, if we initialise with exactly this sharp set of parameters, the training procedure is not able to move away from the initialisation as the gradients associated with the softmax in this region are very small.
Instead, we initialise the controller with a non-ideal version of the generic algorithm.
This means that the choice with the highest probability in the output of the Controller is correct, but the probability of other choices is not zero.
As can be seen in Section~\ref{subsec:anc}, this allows the Controller to learn by gradient descent a new algorithm, different from the original one, that has a lower loss than the ideal version of the compiled program.

\section{Experiments}
\label{sec:exp}
We performed two sets of experiments.
The first shows the capability of the Neural Compiler to perfectly reproduce any given program.
The second shows that our Neural Compiler can adapt and improve the performance of programs.
We present results of data-specific optimisation being carried out and show decreases in runtime for all the algorithms and additionally, for some algorithms, show that the runtime is a different computational-complexity class altogether.
All the code required to reproduce these experiments is available online~\footnote{\texttt{https://github.com/albanD/adaptive-neural-compilation}}.

\subsection{Compilation}
\label{subsec:generalisation-test}
The compiler described in section~\ref{subsec:nc} allows us to go from a program written using our instruction set to a set of weights $\theta$ for our Controller.

To illustrate this point, we implemented simple programs that can solve the tasks introduced by~\citet{nram} and a shortest path problem.
One of these implementations can be found in Figure~\ref{lst:input-program}, while the others are available in appendix~\ref{sec:supp-example-tasks}.
These programs are written in a specific language, and are transformed by the Neural Compiler into parameters for the model.
As expected, the resulting models solve the original tasks exactly and can generalise to any input sequence.

\subsection{ANC experiments}
\label{subsec:anc}
In addition to being able to reproduce any given program as was done by \citet{npi}, we have the possibility of optimising the resulting program further.
We exhibit this by compiling program down to our model and optimising their performance.
The efficiency gains for these tasks come either from finding simpler, equivalent algorithms or by exploiting some bias in the data to either remove instructions or change the underlying algorithm.

We identify three different levels of interpretability for our model:
The first type corresponds to weights containing only Dirac-delta distributions, there is an exact one-to-one mapping between lines in the weight matrices and lines of assembly code.
In the second type where all probabilities are Dirac-delta except the ones associated with the execution of the \texttt{JEZ} instruction, we can recover an exact algorithm that will use \texttt{if} statements to enumerate the different cases arising from this conditional jump.
In the third type where any operation other than \texttt{JEZ} is executed in a soft way or use a soft argument, it is not possible to recover a program that will be as efficient as the learned one.

We present here briefly the considered tasks and biases, and report the reader to appendix~\ref{sec:supp-example-tasks} for a detailed encoding of the input/output tape.
\begin{enumerate}
\item \textbf{Access}: Given a value $k$ and an array $A$, return $A[k]$. In the biased version, the value of $k$ is always be the same, so the address of the required element can be stored in a constant. This is similar to the optimisation known as constant folding.
\item \textbf{Swap}: Given an array $A$ and two pointers $p$ and $q$, swap the elements $A[p]$ and $A[q]$. In the biased version, $p$ and $q$ are always the same so reading them can be avoided.
\item \textbf{Increment}: Given an array, increment all its element by 1. In the biased version, the array is of fixed size and the elements of the array have the same value so you don't need to read all of them when going through the array.
\item \textbf{Listk}: Given a pointer to the head of a linked list, a number $k$ and a linked list, find the value of the $k$-th element. In the biased version, the linked list is organised in order in memory, as would be an array, so the address of the $k$-th value can be computed in constant time. This is the example developed in Figure~\ref{fig:compilation-process}.
\item \textbf{Addition}: Two values are written on the tape and should be summed. No data bias is introduced but the starting algorithm is non-efficient: it performs the addition as a series of increment operation. The more efficient operation would be to add the two numbers.
\item \textbf{Sort}: Given an array $A$, sort it. In the biased version, only the start of the array might be unsorted. Once the start has been arranged, the end of the array can be safely ignored.
\end{enumerate}

For each of these tasks, we perform a grid search on the loss parameters and on our hyper-parameters.
Training is performed using Adam~\citep{kingma2015method}.
We choose the best set of hyperparameters and run the optimisation with 100 different random seeds.
We consider that a program has been successfully optimised when two conditions are fulfilled.
First, it needs to output the correct solution for all test cases presenting the same bias.
Second, the average number of iterations taken to solve a problem must be lower than the algorithm used for initialisation.
Note that if we cared only about the first criterion, the methods presented in Section~\ref{subsec:generalisation-test} would already provide a success rate of 100\%, without requiring any training.

\begin{table}[t]
\captionsetup{font=footnotesize}
\caption[]{Average numbers of iterations required to solve instances of the problems for the original program, the best learned program and the ideal algorithm for the biased dataset. We also include the success rate of reaching a more efficient algorithm across multiple random restarts.}
\label{tab:anc-task-res}
\centering
\begin{tabular}{@{}c@{\hspace*{2ex}}cccccc@{}}
\toprule
& \textbf{Access} & \textbf{Increment} & \textbf{Swap} & \textbf{ListK} & \textbf{Addition} & \textbf{Sort}\\ \hline
\midrule
Generic & 6 & 40 & 10 & 18 & 20 & 38\\ \hline
Learned & 4 & 16 & 6 & 11 & 9 & 18\\ \hline
Ideal & 4 & 34 & 6 & 10 & 6 & 9.5 \\ \hline
\midrule
Success Rate & 37 \% & 84\% & 27\% & 19\% & 12\% & 74\%\\ \hline
\bottomrule
\end{tabular}
\end{table}

The results are presented in Table~\ref{tab:anc-task-res}.
For each of these tasks, we manage to find faster algorithms.
In the simple cases of \textbf{Access} and \textbf{Swap}, the \emph{optimal} algorithm for the presented datasets are obtained.
Exploiting the bias of the data, successful heuristics are incorporated in the algorithm and appropriate constants get stored in the initial value of registers.
The learned programs for these tasks are always in the first case of interpretability, this means that we can recover the most efficient algorithm from the learned weights.

While \textbf{ListK} and \textbf{Addition} have lower success rates, the improvements between the original and learned algorithms are still significant.
Both were initialised with iterative algorithms with $\mathcal{O}(n)$ complexities.
They managed to find constant time $\mathcal{O}(1)$ algorithms to solve the given problems, making the runtime independent of the input.
Achieving this means that the equivalence between the two approaches has been identified, similar to how optimising compilers operate.
Moreover, on the \textbf{ListK} task, some learned programs corresponds to the second type of interpretability.
Indeed these programs use soft jumps to condition the execution on the value of $k$.
Even though these program would not generalise to other values of $k$, some learned programs for this task achieve a type one interpretability and a study of the learned algorithm reveal that they can generalise to \emph{any} value of $k$.

Finally, the \textbf{Increment} task achieves an unexpected result.
Indeed, it is able to outperform our best possible algorithm.
By looking at the learned program, we can see that it is actually leveraging the possibility to perform soft writes over multiple elements of the memory at the same time to reduce its runtime.
This is the only case where we see a learned program associated with the third type of interpretability.
While our ideal algorithm would give a confidence of $1$ on the output, this algorithm is unable to do so, but it has a high enough confidence of $0.9$ to be considered a correct algorithm.

In practice, for all but the most simple tasks, we observe that further optimisation is possible, as some useless instructions remain present.
Some transformations of the controller are indeed difficult to achieve through the local changes operated by the gradient descent algorithm.
An analysis of these failure modes of our algorithm can be found in appendix~\ref{subsec:supp-failure-anal}.
This motivates us to envision the use of approaches other than gradient descent to address these issues.

\section{Discussion}
\label{sec:discuss}
The work presented here is a first step towards adaptive learning of programs.
It opens up several interesting directions of future research.
For exemple, the definition of efficiency that we considered in this paper is flexible.
We chose to only look at the average number of operations executed to generate the output from the input.
We leave the study of other potential measures such as Kolmogorov Complexity and \texttt{sloc}, to name a few, for future works.

As shown in the experiment section, our current method is very good at finding efficient solutions for simple programs.
For more complex programs, only a solution close to the initialisation can be found.
Even though training heuristics could help with the tasks considered here, they would likely not scale up to real applications.
Indeed, the main problem we identified is that the gradient-descent based optimisation is unable to explore the space of programs effectively, by performing only local transformations.
In future work, we want to explore different optimisation methods.
One approach would be to mix global and local exploration to improve the quality of the solutions.
A more ambitious plan would be to leverage the structure of the problem and use techniques from combinatorial optimisation to try and solve the original discrete problem.

\bibliographystyle{plainnat}
\bibliography{bibliography}

\begin{thebibliography}{14}
\providecommand{\natexlab}[1]{#1}
\providecommand{\url}[1]{\texttt{#1}}
\expandafter\ifx\csname urlstyle\endcsname\relax
  \providecommand{\doi}[1]{doi: #1}\else
  \providecommand{\doi}{doi: \begingroup \urlstyle{rm}\Url}\fi

\bibitem[Andrychowicz and Kurach(2016)]{andrychowicz2016learning}
Marcin Andrychowicz and Karol Kurach.
\newblock Learning efficient algorithms with hierarchical attentive memory.
\newblock \emph{arXiv preprint arXiv:1602.03218}, 2016.

\bibitem[Boykov and Kolmogorov(2004)]{efficient-gc}
Yuri Boykov and Vladimir Kolmogorov.
\newblock An experimental comparison of min-cut/max-flow algorithms for energy
  minimization in vision.
\newblock \emph{Pattern Analysis and Machine Intelligence, IEEE Transactions
  on}, 26\penalty0 (9):\penalty0 1124--1137, 2004.

\bibitem[Graves et~al.(2014)Graves, Wayne, and Danihelka]{graves2014neural}
Alex Graves, Greg Wayne, and Ivo Danihelka.
\newblock Neural turing machines.
\newblock \emph{arXiv preprint arXiv:1410.5401}, 2014.

\bibitem[Grefenstette et~al.(2015)Grefenstette, Hermann, Suleyman, and
  Blunsom]{grefenstette2015learning}
Edward Grefenstette, Karl~Moritz Hermann, Mustafa Suleyman, and Phil Blunsom.
\newblock Learning to transduce with unbounded memory.
\newblock In \emph{Advances in Neural Information Processing Systems}, pages
  1819--1827, 2015.

\bibitem[Gruau et~al.(1995)Gruau, Ratajszczak, and Wiber]{gruau1995neural}
Fr{\'e}d{\'e}ric Gruau, Jean-Yves Ratajszczak, and Gilles Wiber.
\newblock A neural compiler.
\newblock \emph{Theoretical Computer Science}, 141\penalty0 (1):\penalty0
  1--52, 1995.

\bibitem[Joulin and Mikolov(2015)]{joulin2015inferring}
Armand Joulin and Tomas Mikolov.
\newblock Inferring algorithmic patterns with stack-augmented recurrent nets.
\newblock In \emph{Advances in Neural Information Processing Systems}, pages
  190--198, 2015.

\bibitem[Kaiser and Sutskever(2016)]{kaiser2015neural}
{\L}ukasz Kaiser and Ilya Sutskever.
\newblock Neural gpus learn algorithms.
\newblock In \emph{ICLR}, 2016.

\bibitem[Kurach et~al.(2016)Kurach, Andrychowicz, and Sutskever]{nram}
Karol Kurach, Marcin Andrychowicz, and Ilya Sutskever.
\newblock Neural random-access machines.
\newblock In \emph{ICLR}, 2016.

\bibitem[Neelakantan et~al.(2016)Neelakantan, Vilnis, Le, Sutskever, Kaiser,
  Kurach, and Martens]{gradient-noise}
Arvind Neelakantan, Luke Vilnis, Quoc~V Le, Ilya Sutskever, Lukasz Kaiser,
  Karol Kurach, and James Martens.
\newblock Adding gradient noise improves learning for very deep networks.
\newblock In \emph{ICLR}, 2016.

\bibitem[Neto et~al.(2003)Neto, Siegelmann, and Costa]{neto2003symbolic}
Jo{\~a}o~Pedro Neto, Hava~T Siegelmann, and J~F{\'e}lix Costa.
\newblock Symbolic processing in neural networks.
\newblock \emph{Journal of the Brazilian Computer Society}, 8\penalty0
  (3):\penalty0 58--70, 2003.

\bibitem[Reed and de~Freitas(2016)]{npi}
Scott Reed and Nando de~Freitas.
\newblock Neural programmer-interpreters.
\newblock In \emph{ICLR}, 2016.

\bibitem[Siegelmann(1994)]{siegelmann1994neural}
Hava~T Siegelmann.
\newblock Neural programming language.
\newblock In \emph{AAAI}, pages 877--882, 1994.

\bibitem[Sutskever et~al.(2014)Sutskever, Vinyals, and
  Le]{sutskever2014sequence}
Ilya Sutskever, Oriol Vinyals, and Quoc~V Le.
\newblock Sequence to sequence learning with neural networks.
\newblock In \emph{Advances in neural information processing systems}, pages
  3104--3112, 2014.

\bibitem[Vinyals et~al.(2015)Vinyals, Fortunato, and
  Jaitly]{vinyals2015pointer}
Oriol Vinyals, Meire Fortunato, and Navdeep Jaitly.
\newblock Pointer networks.
\newblock In \emph{Advances in Neural Information Processing Systems}, pages
  2674--2682, 2015.

\end{thebibliography}


\begin{thebibliography}{18}
\providecommand{\natexlab}[1]{#1}
\providecommand{\url}[1]{\texttt{#1}}
\expandafter\ifx\csname urlstyle\endcsname\relax
  \providecommand{\doi}[1]{doi: #1}\else
  \providecommand{\doi}{doi: \begingroup \urlstyle{rm}\Url}\fi

\bibitem[Andrychowicz and Kurach(2016)]{andrychowicz2016learning}
Marcin Andrychowicz and Karol Kurach.
\newblock Learning efficient algorithms with hierarchical attentive memory.
\newblock \emph{CoRR}, 2016.

\bibitem[Graves et~al.(2014)Graves, Wayne, and Danihelka]{graves2014neural}
Alex Graves, Greg Wayne, and Ivo Danihelka.
\newblock Neural turing machines.
\newblock \emph{CoRR}, 2014.

\bibitem[Grefenstette et~al.(2015)Grefenstette, Hermann, Suleyman, and
  Blunsom]{grefenstette2015learning}
Edward Grefenstette, Karl~Moritz Hermann, Mustafa Suleyman, and Phil Blunsom.
\newblock Learning to transduce with unbounded memory.
\newblock In \emph{NIPS}, 2015.

\bibitem[Gruau et~al.(1995)Gruau, Ratajszczak, and Wiber]{gruau1995neural}
Fr{\'e}d{\'e}ric Gruau, Jean-Yves Ratajszczak, and Gilles Wiber.
\newblock A neural compiler.
\newblock \emph{Theoretical Computer Science}, 1995.

\bibitem[Joulin and Mikolov(2015)]{joulin2015inferring}
Armand Joulin and Tomas Mikolov.
\newblock Inferring algorithmic patterns with stack-augmented recurrent nets.
\newblock In \emph{NIPS}, 2015.

\bibitem[Kaiser and Sutskever(2016)]{kaiser2015neural}
{\L}ukasz Kaiser and Ilya Sutskever.
\newblock Neural gpus learn algorithms.
\newblock In \emph{ICLR}, 2016.

\bibitem[Kingma and Adam(2015)]{kingma2015method}
Diederik Kingma and Jimmy Adam.
\newblock A method for stochastic optimization.
\newblock In \emph{ICLR}, 2015.

\bibitem[Kurach et~al.(2016)Kurach, Andrychowicz, and Sutskever]{nram}
Karol Kurach, Marcin Andrychowicz, and Ilya Sutskever.
\newblock Neural random-access machines.
\newblock In \emph{ICLR}, 2016.

\bibitem[Massalin(1987)]{massalin1987superoptimizer}
Henry Massalin.
\newblock Superoptimizer: a look at the smallest program.
\newblock In \emph{ACM SIGPLAN Notices}, volume~22, pages 122--126. IEEE
  Computer Society Press, 1987.

\bibitem[Neelakantan et~al.(2016)Neelakantan, Vilnis, Le, Sutskever, Kaiser,
  Kurach, and Martens]{gradient-noise}
Arvind Neelakantan, Luke Vilnis, Quoc~V Le, Ilya Sutskever, Lukasz Kaiser,
  Karol Kurach, and James Martens.
\newblock Adding gradient noise improves learning for very deep networks.
\newblock In \emph{ICLR}, 2016.

\bibitem[Neto et~al.(2003)Neto, Siegelmann, and Costa]{neto2003symbolic}
Jo{\~a}o~Pedro Neto, Hava Siegelmann, and F{\'e}lix Costa.
\newblock Symbolic processing in neural networks.
\newblock \emph{Journal of the Brazilian Computer Society}, 2003.

\bibitem[Reed and de~Freitas(2016)]{npi}
Scott Reed and Nando de~Freitas.
\newblock Neural programmer-interpreters.
\newblock In \emph{ICLR}, 2016.

\bibitem[Schkufza et~al.(2013)Schkufza, Sharma, and
  Aiken]{schkufza2013stochastic}
Eric Schkufza, Rahul Sharma, and Alex Aiken.
\newblock Stochastic superoptimization.
\newblock In \emph{ACM SIGARCH Computer Architecture News}, 2013.

\bibitem[Sharma et~al.(2015)Sharma, Schkufza, Churchill, and
  Aiken]{sharma2015conditionally}
Rahul Sharma, Eric Schkufza, Berkeley Churchill, and Alex Aiken.
\newblock Conditionally correct superoptimization.
\newblock In \emph{OOPSLA}, 2015.

\bibitem[Siegelmann(1994)]{siegelmann1994neural}
Hava Siegelmann.
\newblock Neural programming language.
\newblock In \emph{AAAI}, 1994.

\bibitem[Williams(1992)]{williams1992simple}
Ronald Williams.
\newblock Simple statistical gradient-following algorithms for connectionist
  reinforcement learning.
\newblock \emph{Machine learning}, 1992.

\bibitem[Zaremba and Sutskever(2015)]{zaremba2015reinforcement}
Wojciech Zaremba and Ilya Sutskever.
\newblock Reinforcement learning neural turing machines.
\newblock \emph{arXiv preprint arXiv:1505.00521}, 2015.

\bibitem[Zaremba et~al.(2015)Zaremba, Mikolov, Joulin, and
  Fergus]{zaremba2015learning}
Wojciech Zaremba, Tomas Mikolov, Armand Joulin, and Rob Fergus.
\newblock Learning simple algorithms from examples.
\newblock \emph{CoRR}, 2015.

\end{thebibliography}


\begin{thebibliography}{3}
\providecommand{\natexlab}[1]{#1}
\providecommand{\url}[1]{\texttt{#1}}
\expandafter\ifx\csname urlstyle\endcsname\relax
  \providecommand{\doi}[1]{doi: #1}\else
  \providecommand{\doi}{doi: \begingroup \urlstyle{rm}\Url}\fi

\bibitem[Graves et~al.(2014)Graves, Wayne, and Danihelka]{graves2014neural}
Alex Graves, Greg Wayne, and Ivo Danihelka.
\newblock Neural turing machines.
\newblock \emph{CoRR}, 2014.

\bibitem[Grefenstette et~al.(2015)Grefenstette, Hermann, Suleyman, and
  Blunsom]{grefenstette2015learning}
Edward Grefenstette, Karl~Moritz Hermann, Mustafa Suleyman, and Phil Blunsom.
\newblock Learning to transduce with unbounded memory.
\newblock In \emph{NIPS}, 2015.

\bibitem[Kurach et~al.(2016)Kurach, Andrychowicz, and Sutskever]{nram}
Karol Kurach, Marcin Andrychowicz, and Ilya Sutskever.
\newblock Neural random-access machines.
\newblock In \emph{ICLR}, 2016.

\end{thebibliography}

\section{Appendix}
\appendix
\label{sec:supp}
\section{Detailed Model Description}
\label{sec:supp-detailed-model}
In this section, we are going to precisely define the non differentiable  model used above.
This model can be seen as a recurrent network.
Indeed, it takes as input an initial memory tape, performs a certain number of iterations and outputs a final memory tape.
The memory tape is an array of $M$ cells, where a cell is an element holding a single integer value.
The internal state of this recurrent model are the memory, the registers and the instruction register.
The registers are another set of $R$ cells that are internal to the model.
The instruction register is a single cell used in a specific way described later.
These internal states are noted $\mathcal{M}^{t} = \{m_{1}^{t}, m_{2}^{t},\dots, m_{M}^{t}\}$, $\mathcal{R}^{t} = \{r_{1}^{t}, r_{2}^{t},\dots, r_{R}^{t}\}$ and $\mathcal{IR}^{t}$ for the memory, the registers and the instruction register respectively.

Figure 1 describes in more detail how the different elements interact with each other.
At each iteration, the Controller takes as input the value of the instruction register $\mathcal{IR}^{t}$ and outputs four values:
\begin{equation}
  \tt{e}^t, \tt{a}^t, \tt{b}^t, \tt{o}^t = \text{Controller}(\mathcal{IR}^{t}).
\end{equation}
The first value $\tt{e}^t$ is used to select one of the instruction of the Machine to execute at this iteration.
The second and third values $\tt{a}^t$ and $\tt{b}^t$ will identify which registers to use as the first and second argument for the selected instruction.
The fourth value $\tt{o}^t$ identity the output register where to write the result of the executed instruction.
The Machine then takes as input these four values and the internal state and computes the updated value of the internal state and a $stop$ flag:
\begin{equation}
  \mathcal{M}^{t+1}, \mathcal{R}^{t+1}, \mathcal{IR}^{t+1}, stop = \text{Machine}(\mathcal{M}^{t}, \mathcal{R}^{t}, \mathcal{IR}^{t}, \tt{e}^t, \tt{a}^t, \tt{b}^t, \tt{o}^t).
\end{equation}
The $stop$ flag is a binary flag.
When its value is $1$, it means that the model will stop the execution and the current memory state will be returned.

\paragraph{The Machine}
\label{subsec:machine}
The machine is a deterministic function that increments the instruction register and executes the command given by the Controller to update the current internal state.
The set of instructions that can be executed by the Machine can be found in Table~\ref{tab:instr_list}.
Each instruction takes two values as arguments and returns a value.
Additionally, some of these instructions have side effects.
This mean that they do not just output a value, they perform another task.
This other task can be for example to modify the content of the memory.
All the considered side effects can be found in Table~\ref{tab:instr_list}.
By convention, instructions that don't have a value to return and that are used only for their side-effect will return a value of 0.

\paragraph{The Controller}
\label{subsec:controller}
The Controller is a function that takes as input a single value and outputs four different values.
The Controller's internal parameters, the initial values for the registers and the initial value of the instruction register define uniquely a given Controller.

The usual choice in the literature is to use an LSTM network\citep{nram,graves2014neural,grefenstette2015learning} as controller.
Our choice was to instead use a simpler model.
Indeed, our Controller associates a command to each possible value of the instruction register.
Since the instruction register's value will increase by one at each iteration, this will enforce the Controller to encode in its weights what to do at each iteration.
If we were using a recurrent controller the same instruction register could potentially be associated to different sets of outputs and we would lose this one to one mapping.

To make this clearer, we first rewrite the instruction register as an indicator vector with a $1$ at the position of its value:
\begin{equation}
  I_i = \begin{cases}
1 \quad \text{if } i = \mathit{IR}^t\\
0 \quad \text{otherwise}
\end{cases}.
\end{equation}
In this case, we can write a single output $a^t$ of the Controller as the result of a linear function of $I$:
\begin{equation}
  \tt{a}^t = W_{a} * \mathit{I},
  \label{eq:ot}
\end{equation}
where $W_{a}$ is the 1xM matrix containing the value that need to be chosen as first arguments for each possible value of the instruction register and $*$ represent a matrix vector multiplication.

\subsection{Mathematical details of the differentiable model}
\label{subsec:side-eff}
In order to make the model differentiable, every value and every choice are replaced by probability distributions over the possible choices.
Using convex combinations of probability, the execution of the Machine is made differentiable. We present here the mathematical formulation of this procedure for the case of the side-effects.

\paragraph{STOP}
In the discrete model, the execution is halted when the STOP instruction is executed.
However, in the differentiable model, the STOP instruction may be executed with a probability smaller than 1.
To take this into account, when executing the model, we keep track of the probability that the program should have terminated before this iteration based on the probability associated to the STOP instruction at each iteration.
Once this probability goes over a threshold $\eta_{\text{stop}} \in ]0,1]$, the execution is halted.

\paragraph{READ}
The mechanism is entirely the same as the one used to compute the arguments based on the registers and a probability distribution over the registers.

\paragraph{JEZ}
We note $\bm{\mathcal{IR}}^{t+1}_{jez}$ and $\bm{\mathcal{IR}}^{t+1}_{njez}$ the new value of $\bm{\mathcal{IR}}^t$ if we had respectively executed or not the JEZ instruction.
We also have $e^t_{jez}$ the probability of executing this instruction at iteration $t$.
The new value of the instruction register is:
\begin{equation}
  \bm{\mathcal{IR}}^{t+1} = \bm{\mathcal{IR}}^{t+1}_{njez} \cdot (1 - e^t_{jez}) + \bm{\mathcal{IR}}^{t+1}_{jez} \cdot e^t_{jez}
\end{equation}

$\bm{\mathcal{IR}}^{t+1}_{jez}$ is himself computed based on several probability distribution.
If we consider that the instruction JEZ is executed with probabilistic arguments $\mathbf{cond}$ and $\mathbf{label}$, its value is given by
\begin{equation}
\bm{\mathcal{IR}}^{t+1}_{jez} = \mathbf{label} \cdot \text{cond}_{0} + \mathtt{INC}(\bm{\mathcal{IR}}^{t}) \cdot (1 -  \text{cond}_{0})
\end{equation}

With a probability equals to the one that the first argument is equal to zero, the new value of $\bm{\mathcal{IR}}^t$ is $\mathbf{label}$. With the complement, it is equal to the incremented version of its current value, as the machine automatically increments the instruction register.

\paragraph{WRITE}
The mechanism is fairly similar to the one of the JEZ instruction.

We note $\mathbf{M}^{t+1}_{WRITE}$ and $\mathbf{M}^{t+1}_{nWRITE}$ the new value of $\mathbf{M}^{t}$ if we had respectively executed or not the WRITE instruction.
We also have $e^t_{write}$ the probability of executing this instruction at iteration $t$.
The new value of the memory matrix register is:
\begin{equation}
  \mathbf{M}^{t+1} = \mathbf{M}^{t+1}_{nWRITE} \cdot (1 - e^t_{write}) + \mathbf{M}^{t+1}_{WRITE}\cdot e^t_{WRITE}
\end{equation}

As with the JEZ instruction, the value of $\mathbf{M}^{t+1}_{WRITE}$ is dependent on the two probability distribution given as input: $\mathbf{addr}$ and $\mathbf{val}$.
The probability that the $i$-th cell of the memory tape contains the value $j$ after the update is:
\begin{equation}
M_{i,j}^{t+1} = \text{addr}_{i} \cdot \text{val}_{j} + (1 - \text{addr}_{i}) \cdot M_{i,j}^{t}
\end{equation}

Note that this can done using linear algebra operations so as to update everything in one global operation.
\begin{equation}
\mathbf{M}^{t+1} = \left( ((\mathbf{1} - \mathbf{addr}) \mathbf{1}^{T}) \otimes \mathbf{M}^{t} \right)
+ (\mathbf{addr} \ \mathbf{val}^{T})
\end{equation}

\section{Specification of the loss}
\label{sec:supp-loss}
This loss contains four terms that will balance the correctness of the learnt algorithm, proper usage of the stop signal and speed of the algorithms.
The parameters defining the models are the weight of the Controller's function and the initial value of the registers.
When running the model with the parameters $\theta$, we consider that the execution ran for $T$ time steps.
We consider the memory to have a size $M$ and that each number can be an integer between $0$ and $M-1$.
$\mathbf{M}^{t}$ was the state of the memory at the $t$-th step.
$\mathbf{T}$ and $\mathbf{C}$ are the target memory and the 0-1 mask of the elements we want to consider.
All these elements are matrices where for example $\mathbf{M}^{t}_{i,j}$ is the probability of the $i$-th entry of the memory to take the value $j$ at the step $t$.
We also note $p_{\text{stop},t}$ the probability outputted by the Machine that it should have stopped before iteration $t$.

\paragraph{Correctness}
The first term corresponds to the correctness of the given algorithm.
For a given input, we have the expected output and a mask.
The mask allows us to know which elements in the memory we should consider when comparing the solutions.
For the given input, we will compare the values specified by the mask of the expected output with the final memory tape provided by the execution.
We compare them with the $\mathcal{L}_2$ distance in the probability space.
Using the notations from above, we can write this term as:
\begin{equation}
    L_c(\theta) = \sum_{i,j} \mathbf{C}_{i,j} (\mathbf{M}^{T}_{i,j}(\theta) - \mathbf{T}_{i,j})^2.
\end{equation}

If we optimised only this first term, nothing would encourage the learnt algorithm to use the STOP instruction and halt as soon as it finished.

\paragraph{Halting}
To prevent programs to take an infinite amount of time without stopping, we defined a maximum number of iterations $T_{max}$ after which the execution is halted.
During training, we also add a penalty if the Controller didn't halt before this limit:
\begin{equation}
    L_{sT_{max}}(\theta) = (1-p_{\text{stop}-T}(\theta)) \cdot [T == T_{max}]
\end{equation}

\paragraph{Efficiency}
If we consider only the above mentioned losses, the program will make sure to halt by itself but won't do it as early as possible.
We incentivise this behaviour by penalising each iteration taken by the program where it does not stop:
\begin{equation}
    L_t(\theta) = \sum_{t \in [1, T-1]} (1-p_{\text{stop}, t}(\theta)).
\end{equation}

\paragraph{Confidence}
Moreover, we want the algorithm to have a good confidence to stop when it has found the correct output.
To do so, we add the following term which will penalise probability of stopping if the current state of the memory is not the expected one:
\begin{equation}
    L_st(\theta) = \sum_{t \in [2, T]} \sum_{i,j} (p_{\text{stop}, t}(\theta) - p_{\text{stop}, t-1}(\theta)) \mathbf{C}_{i,j} (\mathbf{M}^{t}_{i,j}(\theta) - \mathbf{T}_{i,j})^2.
\end{equation}
The increase in probability $(p_{\text{stop}, t} - p_{\text{stop}, t-1})$ corresponds to the probability of stopping exactly at iteration $t$. So, this is equivalent to the expected error made.

\paragraph{Total loss}
The complete loss that we use is then the following:
\begin{equation}
    L(\theta) = \alpha L_c(\theta) + \beta L_{sT_{max}}(\theta) + \gamma L_st(\theta) + \delta L_t(\theta).
\end{equation}

\section{Distributed representation of the program}
For the most of out experiments, the learned weights are fully interpretable as they fit in the first type of interpretability.
However, in some specific cases, under the pressure of our loss encouraging a smaller number of iterations, an interesting behavior emerges.

\paragraph{Remarks}
It is interesting to note that the decompiled version is not straightforward to interpret.
Indeed when we reach a program that has non Dirac-delta distributions in its weights, we cannot perform the inverse of the one-to-one mapping performed by the compiler.
In fact, it relies on this blurriness to be able to execute the program with a smaller number of instruction.
Notably, by having some blurriness on the \texttt{JEZ} instruction, the program can hide additional instructions, by creating a distributed state.
We now explain the mechanism used to achieve this.

\paragraph{Creating a distributed state}
Consider the following program and assume that the initial value of $\mathcal{IR}$ is $0$:

Initial Registers:\\
$R_{1}=0; R_{2}=1; R_{3}=4, R_{4}=0$\\

Program:\\
$
0: R_{1} = \textbf{READ }(R_{1}, R_{4})\\
1: R_{4} = \textbf{JEZ  }(R_{1}, R_{3})\\
2: R_{4} = \textbf{WRITE}(R_{1}, R_{1})\\
3: R_{4} = \textbf{WRITE}(R_{1}, R_{3})\\
$

If you take this program and execute it for three iterations, it will:
read the first value of the tape into $R_{1}$.
Then, if this value is zero, it will jump to State 4, otherwise it will just increment $\mathcal{IR}$.
This means that depending on the value that was in $R_{1}$, the next instruction that will be executed will be different (in this case, the difference between State 3 and State 4 is which registers they will be writing from).
This is our standard way of implementing conditionals.

Imagine that, after learning, the second instruction in our example program has 0.5 probability of being a \texttt{JEZ} and 0.5 probability of being a \texttt{ZERO}.
If the content of $R_{1}$ is a zero, according to the \texttt{JEZ}, we should jump to State 4, but this instruction is executed with a probability of $0.5$.
We also have $0.5$ probability of executing the \texttt{ZERO} instruction, which would lead to State 3.

Therefore, $\mathcal{IR}$ is not a Dirac-delta distribution anymore but points to State 3 with probability $0.5$ and State 4 with probability $0.5$.

\paragraph{Exploiting a distributed state}
To illustrate, we will discuss how the Controller computes $\mathbf{a}$ for a model with $3$ registers.
The Table \ref{tab:controller-example} show an example of some weights for such a controller.
\begin{table}[h]
  \centering
  \begin{tabular}{l|| c | c | c |}
    & $R_{1}$ & $R_{2}$ & $R_{3}$\\
    \hline
    State 1 & 20 & 5 & -20 \\
    \hline
    State 2 & -20 & 5 & 20 \\
    \hline
  \end{tabular}
  \caption{\label{tab:controller-example} Controller Weights}
\end{table}

If we are in State 1, the output of the controller is going to be
\begin{equation}
  out = \text{softmax}([20, 5, -20]) = [0.9999..., 3e^{-7}, 4e^{-18}]
\end{equation}
If we are in State 2, the output of the controller is going to be
\begin{equation}
  out = \text{softmax}([-20, 5, 20]) = [4e^{-18}, 3e^{-7}, 0.9999...]
\end{equation}

In both cases, the output of the controller is therefore going to be almost discrete. In State 1, $R_{1}$ would be chosen and in State 2, $R_{3}$ would be chosen.

However, in the case where we have a distributed state with probability 0.5 over State 1 and 0.5 over State 2, the output would be:
\begin{equation}
  \begin{split}
    out &= \text{softmax}(0.5 * [-20, 5, 20] + 0.5 [20, 5, -20])\\
    &= \text{softmax}([0, 10 ,0])\\
    &= [4e^{-5}, 0.999, 4e^{-5}].
  \end{split}
\end{equation}
Note that the result of the distributed state is actually different from the result of the discrete states. Moreover it is still a discrete choice of the second register.

Because this program contains distributed elements, it is not possible to perform the one-to-one mapping between the weights and the lines of code.
Though every instruction executed by the program, except for the \texttt{JEZ}, are binary.
This means that this model can be translated to a regular program that will take exactly the same runtime, but will require more lines of codes than the number of lines in the matrix.

\section{Alternative Learning Strategies}
A critique that can be made to this method is that we will still initialise close to a local minimum.
Another approach might be to start from a random initialisation but adding a penalty on the value of the weights such that they are encourage to be close to the generic algorithm.
This can be seen as $\mathcal{L}_2$ regularisation but instead of pushing the weights to $0$, we push then with the value corresponding to the generic algorithm.
If we start with a very high value of this penalty but use an annealing schedule where its importance is very quickly reduced, this is going to be equivalent to the previous method.

\section{Possible Extension}
\subsection{Making objective function differentiable}
These experiments showed that we can transform any program that perform a mapping between an input memory tape to an output memory tape to a set of parameters and execute it using our model.
The first point we want to make here is that this means that we take any program and transform it into a differentiable function easily.
For example, if we want to learn a model that given a graph and two nodes a and b, will output the list of nodes to go through to go from a to b in the shortest amount of time.
We can easily define the loss of the length of the path outputted by the model.
Unfortunately, the function that computes this length from the set of nodes is not differentiable.
Here we could implement this function in our model and use it between the prediction of the model and the loss function to get an end to end trainable system.

\subsection{Beyond mimicking and towards open problems}
It would even be possible to generalise our learning procedure to more complex problems for which we don't have a ground truth output.
For example, we could consider problems where the exact answer for a given input is not computable or not unique.
If the goodness of a solution can be computed easily, this value could be used as training objective.
Any program giving a solution could be used as initialisation and our framework would improve it, making it generate better solutions.

\section{Example tasks}
\label{sec:supp-example-tasks}
This section will present the programs that we use as initialisation for the experiment section.

\subsection{Access}
In this task, the first element in the memory is a value $k$.
Starting from the second element, the memory contains a zero-terminated list.
The goal is to access the $k$-th element in the list that is zero-indexed.
The program associated with this task can be found in Listing~\ref{lst:access}.

{%
\centering
\begin{minipage}{0.25\textwidth}
\begin{lstlisting}[language=neulang, caption= Access Task, label={lst:access}]
var k = 0
k = READ(0)
k = INC(k)
k = READ(k)
WRITE(0, k)
STOP()
\end{lstlisting}
\end{minipage}\par
}

\begin{tabular}{|l|c|c|c|c|c|c|c|c|c|c|c}
\hline
\textbf{Example input:}&6&9&1&2&7&9&8&1&3&5\\
\hline
\textbf{Output:}&1&9&1&2&7&9&8&1&3&5\\
\hline
\end{tabular}

\subsection{Copy}
In this task, the first element in the memory is a pointer $p$.
Starting from the second element, the memory contains a zero-terminated list.
The goal is to copy this list at the given pointer.
The program associated with this task can be found in Listing~\ref{lst:copy}.

{%
\centering
\begin{minipage}{0.6\textwidth}
\begin{lstlisting}[language=neulang, caption= Copy Task, label={lst:copy}]
var read_addr = 0
var read_value = 0
var write_addr = 0

write_addr = READ(0)
l_loop: read_value = READ(read_addr)
JEZ(read_value, l_stop)
WRITE(write_addr, read_value)
read_addr = INC(read_addr)
write_addr = INC(write_addr)
JEZ(0, l_loop)

l_stop: STOP()
\end{lstlisting}
\end{minipage}\par
}
\begin{tabular}{|l|c|c|c|c|c|c|c|c|c|c|c|c|c|c|c|c|}
\hline
\textbf{Example input:}&9&11& 3& 1& 5&14& 0& 0& 0& 0& 0& 0& 0& 0& 0\\
\hline
\textbf{Output:}&9&11& 3& 1& 5&14& 0& 0& 0&11& 3& 1& 5&14& 0\\
\hline
\end{tabular}

\subsection{Increment}
In this task, the memory contains a zero-terminated list.
The goal is to increment each value in the list by $1$.
The program associated with this task can be found in Listing~\ref{lst:increment}.

{%
\centering
\begin{minipage}{0.6\textwidth}
\begin{lstlisting}[language=neulang, caption= Increment Task, label={lst:increment}]
var read_addr = 0
var read_value = 0

l_loop: read_value = READ(read_addr)
JEZ(read_value, l_stop)
read_value = INC(read_value)
WRITE(read_addr, read_value)
read_addr = INC(read_addr)
JEZ(0, l_loop)

l_stop: STOP()
\end{lstlisting}
\end{minipage}\par
}
\begin{tabular}{|l|c|c|c|c|c|c|c|}
\hline
\textbf{Example input:}&1&2&2&3&0&0&0\\
\hline
\textbf{Output:}&2&3&3&4&0&0&0\\
\hline
\end{tabular}

\subsection{Reverse}
In this task, the first element in the memory is a pointer $p$.
Starting from the second element, the memory contains a zero-terminated list.
The goal is to copy this list at the given pointer in the reverse order.
The program associated with this task can be found in Listing~\ref{lst:reverse}.

{%
\centering
\begin{minipage}{0.7\textwidth}
\begin{lstlisting}[language=neulang, caption= Reverse Task, label={lst:reverse}]
var read_addr = 0
var read_value = 0
var write_addr = 0

write_addr = READ(write_addr)
l_count_phase: read_value = READ(read_addr)
JEZ(read_value, l_copy_phase)
read_addr = INC(read_addr)
JEZ(0, l_count_phase)

l_copy_phase: read_addr = DEC(read_addr)
JEZ(read_addr, l_stop)
read_value = READ(read_addr)
WRITE(write_addr, read_value)
write_addr = INC(write_addr)
JEZ(0, l_copy_phase)

l_stop: STOP()
\end{lstlisting}
\end{minipage}\par
}
\begin{tabular}{|l|c|c|c|c|c|c|c|c|c|c|c|c|c|c|c|}
\hline
\textbf{Example input:}&5& 7& 2&13&14& 0& 0& 0& 0& 0& 0& 0& 0& 0& 0\\
\hline
\textbf{Output:}&5&7&2&13&14&14&13& 2& 7& 0& 0& 0& 0& 0& 0\\
\hline
\end{tabular}

\subsection{Permutation}
In this task, the memory contains two zero-terminated list one after the other.
The first contains a set of indices.
the second contains a set of values.
The goal is to fill the first list with the values in the second list at the given index.
The program associated with this task can be found in Listing~\ref{lst:permutation}.

{%
\centering
\begin{minipage}{0.8\textwidth}
\begin{lstlisting}[language=neulang, caption= Permutation Task, label={lst:permutation}]
var read_addr = 0
var read_value = 0
var write_offset = 0

l_count_phase: read_value = READ(write_offset)
write_offset = INC(write_offset)
JEZ(read_value, l_copy_phase)
JEZ(0, l_count_phase)

l_copy_phase: read_value = DEC(read_addr)
JEZ(read_value, l_stop)
read_value = ADD(write_offset, read_value)
read_value = READ(read_value)
WRITE(read_addr, read_value)
read_addr = INC(read_addr)
JEZ(0, l_copy_phase)
l_stop: STOP()
\end{lstlisting}
\end{minipage}\par
}
\begin{tabular}{|l|c|c|c|c|c|c|c|c|c|c|c|c|c|c|c|}
\hline
\textbf{Example input:}&2& 1& 3& 0&13& 4& 6& 0& 0& 0& 0& 0& 0& 0& 0\\
\hline
\textbf{Output:}&4&13& 6& 0&13& 4& 6& 0& 0& 0& 0& 0& 0& 0& 0\\
\hline
\end{tabular}

\subsection{Swap}
In this task, the first two elements in the memory are pointers $p$ and $q$.
Starting from the third element, the memory contains a zero-terminated list.
The goal is to swap the elements pointed by $p$ and $q$ in the list that is zero-indexed.
The program associated with this task can be found in Listing~\ref{lst:swap}.

{%
\centering
\begin{minipage}{0.4\textwidth}
\begin{lstlisting}[language=neulang, caption= Swap Task, label={lst:swap}]
var p = 0
var p_val = 0
var q = 0
var q_val = 0

p = READ(0)
q = READ(1)
p_val = READ(p)
q_val = READ(q)
WRITE(q, p_val)
WRITE(p, q_val)
STOP()
\end{lstlisting}
\end{minipage}\par
}
\begin{tabular}{|l|c|c|c|c|c|c|c|c|c|c|}
\hline
\textbf{Example input:}&1&3&7&6&7&5&2&0&0&0\\
\hline
\textbf{Output:}&1&3&7&5&7&6&2&0&0&0\\
\hline
\end{tabular}

\subsection{ListSearch}
In this task, the first three elements in the memory are a pointer to the head of the linked list, the value we are looking for $v$ and a pointer to a place in memory where to store the result.
The rest of the memory contains the linked list.
Each element in the linked list is two values, the first one is the pointer to the next element, the second is the value contained in this element.
By convention, the last element in the list points to the address $0$.
The goal is to return the pointer to the first element whose value is equal to $v$.
The program associated with this task can be found in Listing~\ref{lst:listsearch}.

{%
\centering
\begin{minipage}{0.7\textwidth}
\begin{lstlisting}[language=neulang, caption= ListSearch Task, label={lst:listsearch}]
var p_out = 0
var p_current = 0
var val_current = 0
var val_searched = 0

val_searched = READ(1)
p_out = READ(2)
l_loop: p_current = READ(p_current)
val_current = INC(p_current)
val_current = READ(val_current)
val_current = SUB(val_current, val_searched)
JEZ(val_current, l_stop)
JEZ(0, l_loop)
l_stop: WRITE(p_out, p_current)
STOP()
\end{lstlisting}
\end{minipage}\par
}
\begin{tabular}{|l|c|c|c|c|c|c|c|c|c|c|c|c|c|c|c|}
\hline
\textbf{Example input:}&11&10& 2& 9& 4& 3&10&0&6&7&13&5&12&0&0\\
\hline
\textbf{Output:}&11&10& 5& 9& 4& 3&10& 0& 6& 7&13& 5&12& 0& 0\\
\hline
\end{tabular}

\subsection{ListK}
In this task, the first three elements in the memory are a pointer to the head of the linked list, the number of hops we want to do $k$ in the list and a pointer to a place in memory where to store the result.
The rest of the memory contains the linked list.
Each element in the linked list is two values, the first one is the pointer to the next element, the second is the value contained in this element.
By convention, the last element in the list points to the address $0$.
The goal is to return the value of the $k$-th element of the linked list.
The program associated with this task can be found in Listing~\ref{lst:listk}.

{%
\centering
\begin{minipage}{0.6\textwidth}
\begin{lstlisting}[language=neulang, caption= ListK Task, label={lst:listk}]
var p_out = 0
var p_current = 0
var val_current = 0
var k = 0

k = READ(1)
p_out = READ(2)
l_loop: p_current = READ(p_current)
k = DEC(k)
JEZ(k, l_stop)
JEZ(0, l_loop)
l_stop: p_current = INC(p_current)
p_current = READ(p_current)
WRITE(p_out, p_current)
STOP()
\end{lstlisting}
\end{minipage}\par
}
\begin{tabular}{|l|c|c|c|c|c|c|c|c|c|c|c|c|c|c|c|c}
\hline
\textbf{Example input:}&3& 2& 2& 9&15& 0& 0& 0& 1&15&17& 7&13& 0& 0&11\\
\hline
\textbf{Output:}&3& 2&17& 9&15& 0& 0& 0& 1&15&17& 7&13& 0& 0&11\\
\hline
\end{tabular}
\begin{tabular}{c|c|c|c|}
\hline
10& 0& 0& 0\\
\hline
10& 0& 0& 0\\
\hline
\end{tabular}

\subsection{Walk BST}
In this task, the first two elements in the memory are a pointer to the head of the BST and a pointer to a place in memory where to store the result.
Starting at the third element, there is a zero-terminated list containing the instructions on how to traverse in the BST.
The rest of the memory contains the BST.
Each element in the BST has three values, the first one is the value of this node, the second is the pointer to the left node and the third is the pointer to the right element.
By convention, the leafs points to the address $0$.
The goal is to return the value of the node we get at after following the instructions.
The instructions are $1$ or $2$ to go respectively to the left or the right.
The program associated with this task can be found in Listing~\ref{lst:walkbst}.

{%
\centering
\begin{minipage}{0.6\textwidth}
\begin{lstlisting}[language=neulang, caption= WalkBST Task, label={lst:walkbst}]
var p_out = 0
var p_current = 0
var p_instr = 0
var instr = 0

p_current = READ(0)
p_out = READ(1)
instr = READ(2)

l_loop: JEZ(instr, l_stop)
p_current = ADD(p_current, instr)
p_current = READ(p_current)
p_instr = INC(p_instr)
JEZ(0, l_loop)

l_stop: p_current = READ(p_current)
WRITE(p_out, p_current)
STOP()
\end{lstlisting}
\end{minipage}\par
}
\begin{tabular}{|l|c|c|c|c|c|c|c|c|c|c|c|c|c|c|c}
\hline
\textbf{Example input:}&12&1&1&2&0&0&15&0&9&23&0&0&11&15&6\\
\hline
\textbf{Output:}&12&10& 1& 2& 0& 0&15& 0& 9&23& 0& 0&11&15& 6\\
\hline
\end{tabular}
\begin{tabular}{c|c|c|c|c|c|c|c|c|c|c|c|c|c|c|c|c|c|c|c|}
\hline
8&0&24&0&0&0&0&0&0&10&0&0&0&0&0\\
\hline
8& 0&24& 0& 0& 0& 0& 0& 0&10& 0& 0& 0& 0& 0\\
\hline
\end{tabular}

\subsection{Merge}
In this task, the first three elements in the memory are pointers to respectively, the first list, the second list and the output.
The two lists are zero-terminated sorted lists.
The goal is to merge the two lists into a single sorted zero-terminated list that starts at the output pointer.
The program associated with this task can be found in Listing~\ref{lst:merge}.

{%
\centering
\begin{minipage}{0.9\textwidth}
\begin{lstlisting}[language=neulang, caption=Merge Task, label={lst:merge}]
var p_first_list = 0
var val_first_list = 0
var p_second_list = 0
var val_second_list = 0
var p_output_list = 0
var min = 0

p_first_list = READ(0)
p_second_list = READ(1)
p_output_list = READ(2)

l_loop: val_first_list = READ(p_first_list)
val_second_list = READ(p_second_list)
JEZ(val_first_list, l_first_finished)
JEZ(val_second_list, l_second_finished)
min = MIN(val_first_list, val_second_list)
min = SUB(val_first_list, min)
JEZ(min, l_first_smaller)

WRITE(p_output_list, val_first_list)
p_output_list = INC(p_output_list)
p_first_list = INC(p_first_list)
JEZ(0, l_loop)

l_first_smaller: WRITE(p_output_list, val_second_list)
p_output_list = INC(p_output_list)
p_second_list = INC(p_second_list)
JEZ(0, l_loop)

l_first_finished: p_first_list = ADD(p_second_list, 0)
val_first_list = ADD(val_second_list, 0)

l_second_finished: WRITE(p_output_list, val_first_list)
p_first_list = INC(p_first_list)
p_output_list = INC(p_output_list)
val_first_list = READ(p_first_list)
JEZ(val_first_list, l_stop)
JEZ(0, l_second_finished)

l_stop: STOP()
\end{lstlisting}
\end{minipage}\par
}

\begin{tabular}{|l|c|c|c|c|c|c|c|c|c|c|c|c|c|c|c}
\hline
\textbf{Example input:}&3& 8&11&27&17&16& 1& 0&29&26& 0& 0& 0& 0& 0 \\
\hline
\textbf{Output:}&3&8&11&27&17&16& 1& 0&29&26& 0&29&27&26&17\\
\hline
\end{tabular}
\begin{tabular}{c|c|c|c|c|c|c|c|c|c|c|c|c|c|c|c|c|c|c|c|}
\hline
 0& 0& 0& 0& 0&0& 0& 0& 0& 0& 0& 0& 0& 0& 0\\
\hline
16& 1& 0& 0& 0& 0& 0& 0& 0& 0& 0& 0& 0& 0& 0\\
\hline
\end{tabular}

\subsection{Dijkstra}
In this task, we are provided with a graph represented in the input memory as follow.
The first element is a pointer $p_out$ indicating where to write the results.
The following elements contain a zero-terminated array with one entry for each vertex in the graph.
Each entry is a pointer to a zero-terminated list that contains a pair of values for each outgoing edge of the considered node.
Each pair of value contains first the index in the first array of the child node and the second value contains the cost of this edge.
The goal is to write a zero-terminated list at the address provided by $p_out$ that will contain the value of the shortest path from the first node in the list to this node.
The program associated with this task can be found in Listings~\ref{lst:dijkstra-1} and \ref{lst:dijkstra-2}.

{%
\centering
\begin{minipage}{0.7\textwidth}
\begin{lstlisting}[language=neulang, caption=Dijkstra Algorithm (Part 1), label={lst:dijkstra-1}]
var min = 0
var argmin = 0

var p_out = 0
var p_out_temp = 0
var p_in = 1
var p_in_temp = 1

var nnodes = 0

var zero = 0
var big = 99

var tmp_node = 0
var tmp_weight = 0
var tmp_current = 0
var tmp = 0

var didsmth = 0

p_out = READ(p_out)
p_out_temp = ADD(p_out, zero)

tmp_current = INC(zero)
l_loop_nnodes:tmp = READ(p_in_temp)
JEZ(tmp, l_found_nnodes)
WRITE(p_out_temp, big)
p_out_temp = INC(p_out_temp)
WRITE(p_out_temp, tmp_current)
p_out_temp = INC(p_out_temp)
p_in_temp = INC(p_in_temp)
nnodes = INC(nnodes)
JEZ(zero, l_loop_nnodes)

l_found_nnodes:WRITE(p_out, zero)
JEZ(zero, l_find_min)
l_min_return:p_in_temp = ADD(p_in, argmin)
p_in_temp = READ(p_in_temp)

l_loop_sons:tmp_node = READ(p_in_temp)
JEZ(tmp_node, l_find_min)
tmp_node = DEC(tmp_node)
p_in_temp = INC(p_in_temp)
tmp_weight = READ(p_in_temp)
p_in_temp = INC(p_in_temp)

p_out_temp = ADD(p_out, tmp_node)
p_out_temp = ADD(p_out_temp, tmp_node)
tmp_current = READ(p_out_temp)
tmp_weight = ADD(min, tmp_weight)

tmp = MIN(tmp_current, tmp_weight)
tmp = SUB(tmp_current, tmp)
JEZ(tmp, l_loop_sons)
WRITE(p_out_temp, tmp_weight)
JEZ(zero, l_loop_sons)

\end{lstlisting}
\end{minipage}\par}
{%
\centering
\begin{minipage}{0.7\textwidth}
\begin{lstlisting}[language=neulang, caption=Dijkstra Algorithm (Part 2), label={lst:dijkstra-2}, firstnumber=57]
l_find_min:p_out_temp = DEC(p_out)
tmp_node = DEC(zero)
min = ADD(big, zero)
argmin = DEC(zero)

l_loop_min:p_out_temp = INC(p_out_temp)
tmp_node = INC(tmp_node)
tmp = SUB(tmp_node, nnodes)
JEZ(tmp, l_min_found)

tmp_weight = READ(p_out_temp)

p_out_temp = INC(p_out_temp)
tmp = READ(p_out_temp)
JEZ(tmp, l_loop_min)

tmp = MAX(min, tmp_weight)
tmp = SUB(tmp, tmp_weight)
JEZ(tmp, l_loop_min)
min = ADD(tmp_weight, zero)
argmin = ADD(tmp_node, zero)
JEZ(zero, l_loop_min)

l_min_found:tmp = SUB(min, big)
JEZ(tmp, l_stop)
p_out_temp = ADD(p_out, argmin)
p_out_temp = ADD(p_out_temp, argmin)
p_out_temp = INC(p_out_temp)
WRITE(p_out_temp, zero)
JEZ(zero, l_min_return)

l_stop:STOP()
\end{lstlisting}
\end{minipage}\par
}
\textit{Example omitted for space reasons}

\section{Learned optimisation: Case study}
Here we present an analysis of the optimisation achieved by the ANC.
We take the example of the \textbf{ListK} task and study the difference between the learned program and the initialisation used.

\subsection{Representation} The representation chosen is under the form of the intermediary representation described in Figure (2b).
Based on the parameters of the Controller, we can recover the approximate representation described in Figure (2b):
1For each possible "discrete state" of the instruction register, we can compute the commands outputted by the controller.
We report the most probable value for each distribution, as well as the probability that the compiler would assign to this value.
If no value has a probability higher than 0.5, we only report a neutral token (\texttt{R-}, \texttt{-}, \texttt{NOP}).

\subsection{Biased ListK}Figure~\ref{fig:ref-prog} represents the program that was used as initialisation to the optimisation problem.
This is the direct result from the compilation performed by the Neural Compiler of the program described in Listing~\ref{lst:listk}.
A version with a probability of 1 for all necessary instructions would have been easily obtained but not amenable to learning.

Figure~\ref{fig:ref-learned} similarly describes the program that was obtained after learning.

As a remainder, the bias introduced in the ListK task is that the linked list is well organised in memory.
In the general case, the element could be in any order.
An input memory tape to the problem of asking for the third element in the linked list containing $\{4,5,6,7\}$ would be:

\begin{tabular}{|c|c|c|c|c|c|c|c|c|c|c|c|c|c|c|c|c|c|c|c|}
\hline
9& 3& 2& 0& 0&11& 5& 0& 7& 5& 4& 7& 6& 0& 0& 0& 0& 0& 0& 0\\
\hline
\end{tabular}

or

\begin{tabular}{|c|c|c|c|c|c|c|c|c|c|c|c|c|c|c|c|c|c|c|c|}
\hline
5& 3& 2& 0& 0& 7& 4&15& 5& 0& 7& 0& 0& 0& 0& 9& 6& 0& 0& 0\\
\hline
\end{tabular}

In the biased version of the task, all the elements are arranged in order and contiguously positioned on the tape.
The only valid representation of this problems is:

\begin{tabular}{|c|c|c|c|c|c|c|c|c|c|c|c|c|c|c|c|c|c|c|c|}
\hline
3& 3& 2& 5& 4& 7& 5& 9& 6&0& 7& 0& 0& 0& 0& 0& 0& 0& 0& 0\\
\hline
\end{tabular}\\

\subsection{Solutions}Because of the additional structure of the problem, the bias in the data, a more efficient algorithm to find the solution exists.
Let us dive into the comparison of the two different solutions.

Both use their first two states to read the parameters of the given instance of the task.
Which element of the list should be returned is read at line (\texttt{0:}) and where to write the returned value is read at line (\texttt{1:}).
Step (\texttt{2:}) to (\texttt{6:}) are dedicated to putting the address of the $k$-th value of the linked list into the registers \texttt{R1}.
Step (\texttt{7:}) to (\texttt{9:}) perform the same task in both solution: reading the value at the address contained in \texttt{R1}, writing it at the desired position and stopping.

The difference between the two programs lies in how they put the address of the $k$-th value into \texttt{R1}.

\paragraph{Generic} The initialisation program, used for initialisation, works in the most general case so needs to perform a loop where it put the address of the next element in the linked list in \texttt{R1} (\texttt{2:}), decrement the number of jumps remaining to be made (\texttt{3:}), checking whether the wanted element has been reached (\texttt{4:}) and going back to the start of the loop if not (\texttt{5:}).
Once the desired element is reached, \texttt{R1} is incremented so as to point on the value of the linked list element.

\paragraph{Specific} On the other hand, in the biased version of the problem, the position of the desired value can be analytically determined. The function parameters occupy the first three cells of the tape.
After those, each element of the linked list will occupy two cells (one for the pointer to the next address and one for the value).
Therefore, the address of the desired value is given by
\begin{equation}
\begin{split}
\mathtt{R1} &= 3 + (2 * (k-1) + 1) - 1 + 1\\
& = 3 + 2*k-1
\end{split}
\end{equation}
(the -1 comes from the fact that the address are 0-indexed and the final +1 from the fact that we are interested in the position of the value and not of the pointer.)

The way this is computed is as follows:
\begin{itemize}
\renewcommand\labelitemi{-}
\item $\mathtt{R1} = 3 + k$ by adding the constant 3 to the registers \texttt{R2} containing K. \hfill(\texttt{2:})
\item $\mathtt{R2} = k - 1$  \hfill(\texttt{3:})
\item $\mathtt{R1} = 3 + 2*k - 1$ by adding the now reduced value of \texttt{R2}. \hfill(\texttt{6:})
\end{itemize}

The algorithm implemented by the learned version is therefore much more efficient for the biased dataset, due to its capability to ignore the loop.

\subsection{Failure analysis}
\label{subsec:supp-failure-anal}
An observation that can be made is that in the learned version of the program, Step (\texttt{4:}) and (\texttt{5:}) are not contributing to the algorithms.
They execute instructions that have no side effect and store the results into the registers \texttt{R7} that is never used later in the execution.

The learned algorithm could easily be more efficient by not performing these two operations.
However, such an optimisation, while perhaps trivial for a standard compiler, capable of detecting unused values, is fairly hard for our optimisers to discover.
Because we are only doing gradient descent, the action of "moving some instructions earlier in the program" which would be needed here to make the useless instructions disappear, is fairly hard, as it involves modifying several rows of the program at once in a coherent manner.

\begin{figure}[h]
\begin{framed}
\small
\begin{verbatim}
R1 = 0 (0.88)
R2 = 1 (0.88)
R3 = 2 (0.88)
R4 = 6 (0.88)
R5 = 0 (0.88)
R6 = 2 (0.88)
R7 = - (0.05)

Initial State: 0 (0.88)

0: 	R2 (0.96)  	= READ (0.93)  	[ R2 (0.96)  	, R- (0.14)  	]
1: 	R3 (0.96)  	= READ (0.93)  	[ R3 (0.96)  	, R- (0.14)  	]
2: 	R1 (0.96)  	= READ (0.93)  	[ R1 (0.96)  	, R- (0.14)  	]
3: 	R2 (0.96)  	= DEC  (0.93)  	[ R2 (0.96)  	, R- (0.14)  	]
4: 	R7 (0.96)  	= JEZ  (0.93)  	[ R2 (0.96)  	, R4 (0.96)  	]
5: 	R7 (0.96)  	= JEZ  (0.93)  	[ R5 (0.96)  	, R6 (0.96)  	]
6: 	R1 (0.96)  	= INC  (0.93)  	[ R1 (0.96)  	, R- (0.14)  	]
7: 	R1 (0.96)  	= READ (0.93)  	[ R1 (0.96)  	, R- (0.14)  	]
8: 	R7 (0.96)  	= WRIT (0.93)  	[ R3 (0.96)  	, R1 (0.96)  	]
9: 	R7 (0.96)  	= STOP (0.93)  	[ R- (0.14)  	, R- (0.14)  	]
10: 	R- (0.16)  	= NOP  (0.11)  	[ R- (0.16)  	, R- (0.16)  	]
11: 	R- (0.16)  	= NOP  (0.11)  	[ R- (0.18)  	, R- (0.16)  	]
12: 	R- (0.17)  	= NOP  (0.1)  	[ R- (0.17)  	, R- (0.17)  	]
13: 	R- (0.16)  	= NOP  (0.11)  	[ R- (0.17)  	, R- (0.16)  	]
14: 	R- (0.17)  	= NOP  (0.11)  	[ R- (0.16)  	, R- (0.16)  	]
15: 	R- (0.16)  	= NOP  (0.1)  	[ R- (0.16)  	, R- (0.17)  	]
16: 	R- (0.17)  	= NOP  (0.11)  	[ R- (0.16)  	, R- (0.18)  	]
17: 	R- (0.18)  	= NOP  (0.1)  	[ R- (0.18)  	, R- (0.17)  	]
18: 	R- (0.17)  	= NOP  (0.1)  	[ R- (0.16)  	, R- (0.16)  	]
19: 	R- (0.15)  	= NOP  (0.11)  	[ R- (0.16)  	, R- (0.17)  	]
\end{verbatim}
\end{framed}
\caption{\label{fig:ref-prog}Initialisation used for the learning of the ListK task.}
\end{figure}

\begin{figure}[h]
\begin{framed}
\small
\begin{verbatim}
R1 = 3 (0.99)
R2 = 1 (0.99)
R3 = 2 (0.99)
R4 = 10 (0.53)
R5 = 0 (0.99)
R6 = 2 (0.99)
R7 = 7 (0.99)

Initial State: 0 (0.99)

0: 	R2 (0.99)  	= READ (0.99)  	[ R2 (1)  	, R1 (0.97)  	]
1: 	R6 (0.99)  	= READ (0.99)  	[ R3 (0.99)  	, R6 (0.5)  	]
2: 	R1 (1)  	= ADD  (0.99)  	[ R1 (0.99)  	, R2 (1)  	]
3: 	R2 (1)  	= DEC  (0.99)  	[ R2 (1)  	, R1 (0.99)  	]
4: 	R7 (0.99)  	= MAX  (0.99)  	[ R2 (0.99)  	, R1 (0.51)  	]
5: 	R7 (0.99)  	= INC  (0.99)  	[ R6 (0.7)  	, R1 (0.89)  	]
6: 	R1 (1)  	= ADD  (0.99)  	[ R1 (0.99)  	, R2 (1)  	]
7: 	R1 (0.99)  	= READ (0.99)  	[ R1 (0.99)  	, R1 (0.53)  	]
8: 	R7 (0.99)  	= WRIT (0.99)  	[ R6 (1)  	, R1 (0.99)  	]
9: 	R7 (0.9)  	= STOP (0.99)  	[ R6 (0.98)  	, R1 (0.99)  	]
10: 	R2 (0.99)  	= STOP (0.96)  	[ R1 (0.52)  	, R1 (0.99)  	]
11: 	R1 (0.98)  	= ADD  (0.73)  	[ R4 (0.99)  	, R2 (0.99)  	]
12: 	R3 (0.98)  	= ADD  (0.64)  	[ R6 (0.99)  	, R1 (0.99)  	]
13: 	R3 (0.87)  	= STOP (0.65)  	[ R3 (0.52)  	, R1 (0.99)  	]
14: 	R3 (0.89)  	= STOP (0.62)  	[ R6 (0.99)  	, R2 (0.62)  	]
15: 	R3 (0.99)  	= STOP (0.65)  	[ R3 (0.99)  	, R2 (0.71)  	]
16: 	R3 (0.99)  	= NOP  (0.45)  	[ R6 (0.99)  	, R1 (0.99)  	]
17: 	R2 (0.99)  	= INC  (0.56)  	[ R6 (0.7)  	, R1 (0.98)  	]
18: 	R3 (0.99)  	= STOP (0.65)  	[ R3 (0.99)  	, R1 (0.99)  	]
19: 	R3 (0.98)  	= STOP (0.98)  	[ R2 (0.62)  	, R1 (0.79)  	]
\end{verbatim}
\caption{\label{fig:ref-learned} Learnt program for the listK task}
\end{framed}
\end{figure}

\end{document}